\documentclass[letterpaper]{article} %
\usepackage[]{aaai25}  %
\nocopyright
\usepackage{times}  %
\usepackage{helvet}  %
\usepackage{courier}  %
\usepackage[hyphens]{url}  %
\usepackage{graphicx} %
\usepackage{natbib}  %
\usepackage{caption} %
\usepackage{algorithm}
\usepackage{algorithmic}

\usepackage{amsmath}
\usepackage{makecell}
\usepackage{booktabs,multirow}
\usepackage{subcaption}
\usepackage{tabularx}
\usepackage{newfloat}
\usepackage{listings}
\DeclareCaptionStyle{ruled}{labelfont=normalfont,labelsep=colon,strut=off} %
\floatstyle{ruled}
\newfloat{listing}{tb}{lst}{}
\floatname{listing}{Listing}
\title{Enhancing Multimodal Affective Analysis with Learned Live Comment Features}
\author{
    Zhaoyuan Deng,
    Amith Ananthram,
    Kathleen McKeown
}
\affiliations{
    Columbia University\\

    zd2286@columbia.edu,
    amith.ananthram@columbia.edu,
    kathy@cs.columbia.edu
}

\usepackage{bibentry}

\begin{document}

\maketitle

\begin{abstract}
Live comments,
also known as {\em Danmaku}, are user-generated messages that are synchronized with video content. %
These comments overlay directly onto streaming videos, %
capturing viewer emotions and reactions in real-time. While prior %
work has 
leveraged live comments in affective analysis, %
its use has been limited due to the relative rarity %
of live comments across different video platforms. To address this, we first construct %
the Live Comment for Affective Analysis (LCAffect) dataset which contains live comments for English and Chinese videos spanning diverse genres that %
elicit a wide spectrum of emotions. %
Then, using this dataset, we use contrastive learning to train a video encoder %
to produce synthetic %
live comment features for enhanced %
multimodal affective content analysis. Through %
comprehensive experimentation on a wide range of affective analysis tasks (sentiment, emotion recognition, and sarcasm detection) in both English and Chinese, %
we demonstrate that these synthetic live comment features %
significantly improve performance over state-of-the-art methods. We will release our dataset, code, and models upon publication.%
\end{abstract}

\section{Introduction}

\begin{figure}[t]
  \includegraphics[width=\columnwidth]{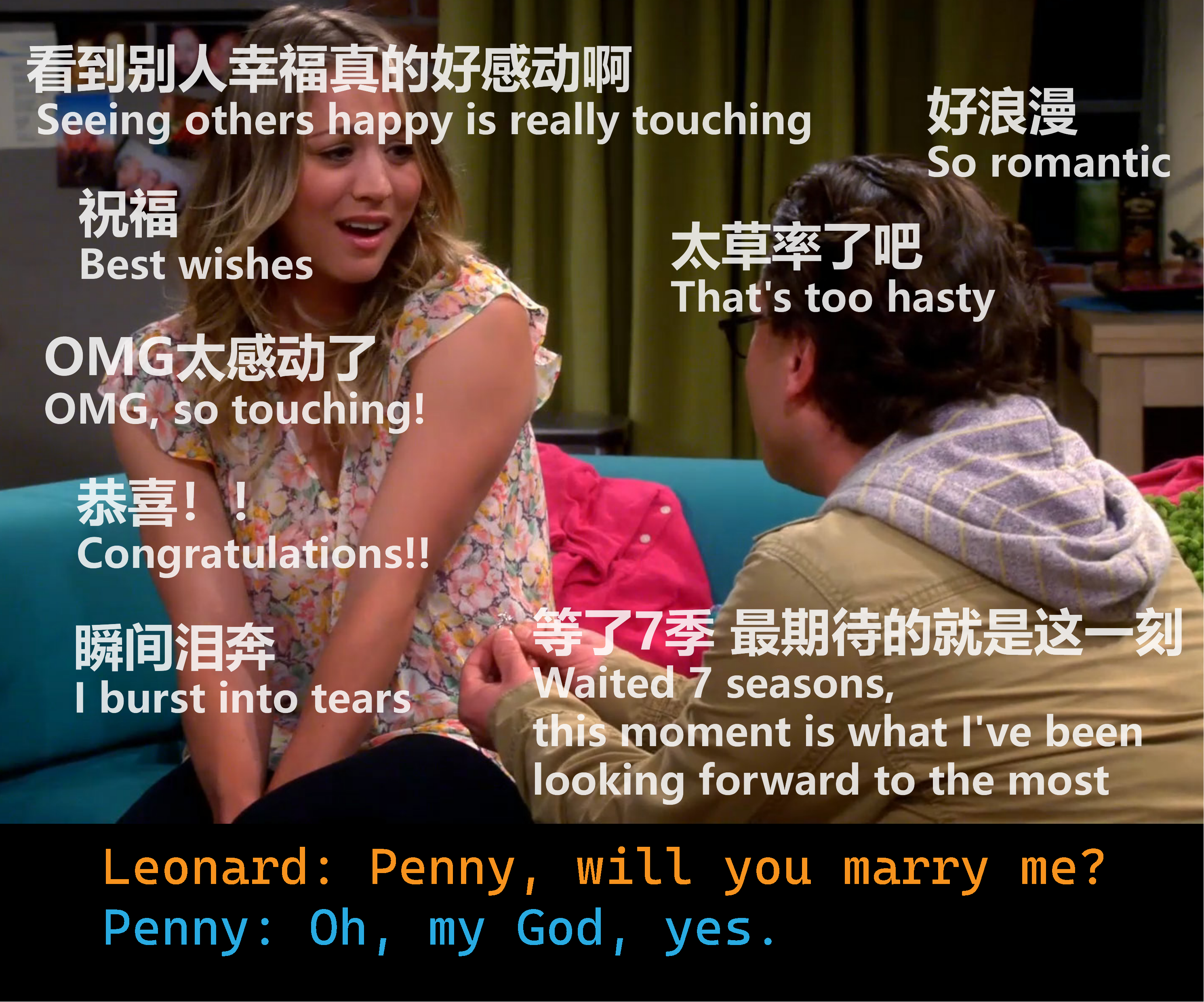}
  \caption{%
  An example video frame and transcript from the TV show \textit{The Big Bang Theory} with accompanying live comments overlaid. Below each original Chinese comment, we include its English translation. By training a video encoder to produce representations similar to these live comments, we can produce multimodal features that adapt well to affective analysis tasks like emotion recognition.%
  }
  \label{fig:experiments}
\end{figure}

Live comments%
, or \textit{Danmaku}, are a subtitle %
system originating from Japan and popularized in China. This system overlays real-time user-generated messages directly onto online video streams. These comments, synchronized with a video's timeline, offer a rich layer of viewer interaction, providing insights into user emotions, user reactions, and the video's broader context %
(see Figure \ref{fig:experiments}). %
The richness of these comments makes them an ideal resource for enhancing multimodal affective analysis, which seeks to interpret and categorize the emotional content and dynamics of videos. 
Although prior research, such as the study by \citet{8560134}, has demonstrated how live comments can %
improve performance in affective analysis tasks  like emotion prediction, %
their broader application remains constrained. Many video platforms do not support live comments, rendering these insights %
impossible for videos on those platforms. Consequently, the %
rarity of live comments poses significant challenges in fully exploiting their analytical potential. %

In this work, we bridge this gap by proposing a novel %
method for producing synthetic live comment features to improve the performance of multimodal affective analysis. We demonstrate the effectiveness of these features in English and Chinese across %
several %
video understanding tasks: %
sentiment analysis, emotion recognition, and sarcasm detection.

One major challenge in affective learning from live comments stems from the nature of existing large-scale live comment datasets which are primarily tailored to the live comment generation task. %
These datasets typically feature a %
large percentage of popular content genres such as video games, sports, and music videos. However, these categories typically lack the rich emotional context required for detailed affective analysis. This mismatch complicates the training of models capable of accurately understanding and interpreting nuanced emotions. %
Additionally, the majority of these datasets are collected from Bilibili, which features amateur-created videos of everyday life. These videos often provoke a more limited and homogeneous range of emotional responses \citep{doi:10.1177/1948550618820309}%
, especially when compared to the diverse and intense emotions elicited by content %
like dramas or documentaries. Moreover, most datasets comprise exclusively Chinese content, which further restricts their multilingual applicability. %

A second significant challenge is the dependency of existing methods on the availability of live comments. In the real-world, the presence of live comments can be %
limited and not uniformly distributed across all video genres. This reliance restricts the practical application of these methods, as they cannot work in scenarios where comments are non-existent.

To address these challenges, we first construct a large-scale, diverse dataset comprising %
everyday life videos in Chinese %
as well as bilingual TV shows, movies, and documentaries, enabling our system to learn a broad spectrum of emotional experiences across various contexts.
Then, utilizing our dataset, we develop a multimodal encoder that effectively learns from %
our collected videos and comments by contrasting different contexts. This encoder is capable of producing synthetic live comment features from videos, thereby enabling the inference of emotional context across modalities, even in the absence of live comments. %

Our experimental analysis demonstrates that adding our live comment features improves over state-of-the-art (SOTA) approaches that leverage text, acoustic, and visual modalities directly. Our system achieves new SOTA performance on multiple multimodal datasets, particularly enhancing accuracy in detecting sentiment, emotions, and sarcasm. %
This not only underscores the utility of live comments in affective video analysis but also opens new avenues for multimodal learning applications.

Our main contributions are: %
\begin{itemize}
\item1) %
Live Comment for Affective Analysis dataset (LCAffect), an expansive corpus of over $11$ million live comments. %
It includes bilingual video content (Chinese and English) spanning a diverse range of topics and genres, curated to capture a wide spectrum of emotions. %
\item2) A contrastive encoder that learns to project videos into live comment representation space, allowing the inference of synthetic live comment features for any video. %
\item3) A downstream multimodal fusion model for affective analysis tasks which utilizes not only three modalities but also our synthetic %
live comment feature. 
\item%
4) New SOTA across three affective analysis tasks, including   
a $3.18$-point increase in accuracy on CH-SIMS v2 (sentiment analysis), a $2.89$-point increase in F1 on MELD (emotion recognition), and a $3.0$-point increase in F1 on MuSTARD (sarcasm detection).%
\end{itemize}

\section{Related Work}
\noindent\textbf{Learning from Comments}\quad
Previous research has highlighted the value of user comments in improving performance across various multimodal %
tasks. \citet{fu-etal-2017-video} show that live comments on Twitch help predict highlights in e-sport games. Similarly, \citet{8695340} show that live comments help identify humor in online videos. Additionally, \citet{8560134} show that features derived from live comments improve %
affective content analysis. Moreover, \citet{hanu2022vtc} show that user comments can help learn more contextualized representations for image, video, and audio, thus improving video-text retrieval. 

These studies depend on the availability of user comments which are not always present in the real-world. In contrast, our work seeks to develop a system that is pre-trained to align video segments with corresponding live comments. This approach aims to learn a multimodal representation space that effectively augments various downstream understanding tasks. Consequently, our system can be directly applied to videos lacking associated live comments, producing synthetic live comment features for such videos. This capability directly addresses the limitations of prior work, enhancing the analysis and interpretation of videos that would be excluded due to the absence of live comments.

\noindent\textbf{Live Comment Datasets}\quad
Several datasets featuring live comments have been established in the literature. LiveBot \citep{Ma2018LiveBotGL} and VideoIC \cite{10.1145/3394171.3413890} are %
large-scale datasets drawn from Bilibili, a site which primarily features short, user-generated videos focused on everyday life. In contrast, \citet{Lalanne2023LiveChatVC} collected live-streamed video and comments from Twitch, a platform centered around video games. Finally, MovieLC \citep{10219855} is a compilation of famous movies with accompanying comments from Tencent Video.

Our work builds on these efforts by creating a new dataset that includes videos from a broader range of platforms. %
By integrating multiple video hosting platforms, our dataset includes %
both a variety of short, user-generated content %
and longer formats like TV shows, movies, and documentaries in Chinese and English. This %
expansive collection allows the study of many genres and 
enables the training of a multimodal encoder that can learn %
live comment features effectively for 
affective analysis %
from diverse videos. %

\noindent\textbf{Multimodal Affective Analysis
}\quad
Previously, multimodal affective computing often relied on hand-crafted algorithms to perform initial feature extraction. Recently, however, there has been a shift towards using pre-trained modality encoders with %
end-to-end tuning %
to improve performance. Specifically, \citet{YI2024111136} use CLIP \citep{radford2021learningtransferablevisualmodels}'s spatial encoder and TimeSformer \citep{pmlr-v139-bertasius21a} to encode visual features, while \citet{wu2024multimodal} employ pre-trained audio encoders such as Data2Vec \citep{baevski2022data2vecgeneralframeworkselfsupervised} and Hubert \citep{10.1109/TASLP.2021.3122291} to process acoustic features. Although these studies demonstrate the efficacy of an end-to-end framework, they are limited to two modalities. Our work extends this approach by integrating all three modalities (text, acoustic, visual), achieving more %
robust video analysis. Another promising research direction in affective analysis involves the integration of external knowledge. \citet{ghosal-etal-2020-cosmic} employ features produced by a common sense knowledge model to enhance emotion recognition, while \citet{hu-etal-2022-unimse} exploit the complementary knowledge underlying sentiment analysis and emotion recognition to build a knowledge-sharing framework. Our work parallels these efforts by leveraging external knowledge derived from live comments. %

\section{The Live Comment Dataset}
\subsection{Collection of Videos}
We collect videos and their accompanying live comments from several popular Chinese video streaming websites: Bilibili, Tencent Video, iQIYI, and Youku. Our dataset includes both Chinese and English content and is %
partitioned into three distinct subsets based on content type: %

\noindent\textbf{User-Generated Content
}\quad This subset includes $2,464$ videos from Bilibili featuring naturally occurring conversations from %
everyday life. %
The initial set of $682$ videos is available from the Linguistic Data Consortium (UPenn) and was manually inspected to ensure content quality and relevance. Each video includes interactions involving at least two people with observable emotions. %
The remaining $1,782$ videos are sourced from Bilibili's recommendation API, which selects content similar to the manually inspected set. 

\noindent\textbf{TV shows and documentaries
}\quad We collect $443$ episodes from $8$ TV shows and documentaries. This subset includes %
content from multiple platforms, %
including dramas, sitcoms, and documentaries focused on social life. %
Their inclusion aims to provide the model with insights %
into scripted content that elicits a wide range of emotions in different contexts.

\noindent\textbf{Movies}\quad Sourced from Tencent Video, this subset features $59$ popular comedy, drama, and action films. They feature the longest videos, %
enabling the analysis of viewer interactions over extended contexts through live comments.%

In addition, we %
collect text transcripts for the videos. For user-generated content, transcripts are obtained using Bilibili's API, which uses automatic speech recognition. For movies and TV shows, we extract transcripts from hard-coded subtitles via optical character recognition (OCR). We extract both English and Chinese subtitles when %
available. %

\subsection{Dataset Statistics}
Our LCAffect dataset contains $11,356,212$ live comments aligned with $829$ hours of video. Table \ref{tab:dataset_stats} provides a detailed breakdown of the dataset, including subset-specific statistics. As shown in Table \ref{tab:dataset_compare}, our dataset establishes a new standard in %
live comment analysis by offering nearly twice the video duration found in the largest existing datasets, encompassing a broader variety of content types, and featuring over twice %
as many live comments %
as any comparable dataset.

\begin{table}

  \centering
  \begin{tabular}{lccc}
    \toprule
    \textbf{Category}           & \textbf{\makecell{User G \\ Content}}           & \textbf{\makecell{TV\\Shows}} & \textbf{Movies} \\
    \cmidrule(r){1-1}\cmidrule(r){2-4}
    \# Videos & 2,464  & 443 & 59                    \\
    \# w. Eng. Subtitle & -  & 281 & 33                    \\
    \# Comm. (k)      & 2,367         &           4,272      &  4,717         \\
    Dur. (h)      &    493.4       & 212.7 & 122.9                          \\
    \cmidrule(r){1-1}\cmidrule(r){2-4}
    Avg. Dur. (s)      &   720.9         &  1,728.5  & 7,500.0                       \\
    Avg. \# Comm      &   960.6       & 9,643.4 & 79,956.6                         \\
    Avg. \# Char      &    8.8       & 12.1 &     9.9                                         \\
    \bottomrule
  \end{tabular}
  \caption{
    Subset-specific statistics of our LCAffect dataset.
  }%
  \label{tab:dataset_stats}
\end{table}

\begin{table}
  \setlength{\tabcolsep}{1mm}
  \centering
  \begin{tabular}{lcccc}
    \toprule
    \textbf{Dataset}           & \textbf{LiveBot}           & \textbf{VideoIC} & \textbf{MovieLC} & \textbf{LCAffect} \\
    \cmidrule(r){1-1}\cmidrule(r){2-5}
    \# Videos & 2,361  & 4,951 & 85 & 2,966                  \\
    
    \# Comm. (k)& 895 & 5,330 & 1,406 & 11,356 \\
    Dur. (h)  & 114   &557  &175&829
                        \\
    Content Lang. &zh &zh&zh&zh \& en\\
    Composition &\makecell{User G.\\Content}&\makecell{User G.\\Content}&Mov.&\makecell{User G.C.\\TV, Mov.} \\
    \bottomrule
  \end{tabular}
  \caption{
    Comparison with other live comment datasets.
  }
  \label{tab:dataset_compare}
\end{table}
\section{Contrastive Pre-training}
Although multiple studies have demonstrated the effectiveness of live comments %
in enhancing video understanding, the availability of live comments %
is largely limited outside of China and Japan. This restriction significantly limits the practical application of these studies. To overcome this challenge, we propose a %
video encoder pre-trained on our diverse live comment dataset. This model leverages a standard contrastive learning framework akin to CLIP \citep{radford2021learningtransferablevisualmodels}, where positive examples consist of matching live comments and videos, and negative examples involve non-matching pairs. The aim is to produce a synthetic live comment feature from a video span %
that can be used to enhance performance in various affective analysis tasks. %
\begin{figure*}[t]
   \centering
  \includegraphics[width=0.98\linewidth]{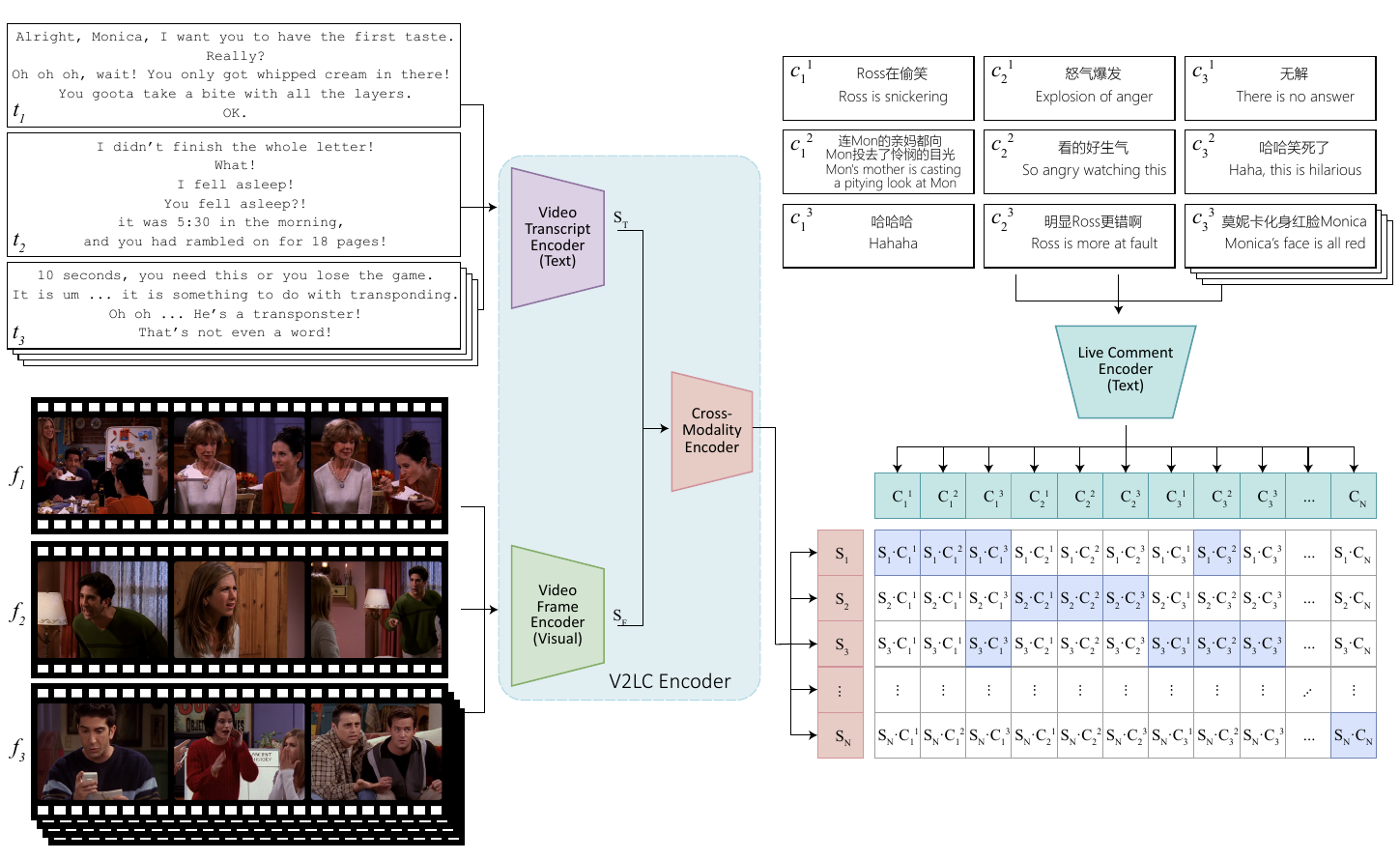} \hfill
 
  \caption {%
  Our contrastive pre-training approach. We train our V2LC encoder to predict the correct pairings of a batch of (video, live comment) training examples. Correct pairings of video segment embedding $S$ and live comment embedding $C$ are highlighted in blue. Comments ${c_1}^3$ and ${c_3}^2$ have high similarity; thus they are correct matches for both Segment 1 and 3.}
\end{figure*}

\subsection{Pre-training Model Architecture}
We %
adopt a CLIP-style contrastive pre-training approach. During the training phase, we partition each video into segments $s$ of $\sigma$ seconds. For each segment $s_i$, we gather the associated text transcript $t_i$, video frames $f_i$, and live comments $c_i$, and treat the segment as an individual training sample. %
For this stage, we use only the text and visual modalities. We create our Video-to-Live Comment  (V2LC) encoder %
by employing a pre-trained text encoder to encode the video text transcript into embedding $\mathbf{S}_{\mathrm{T}}$ %
and a Video Vision Transformer \citep{9710415} to encode the video frames into embedding $\mathbf{S}_{\mathrm{F}}$. %
These encoded outputs are then integrated using a cross-modality encoder to produce the final output video segment embedding $\mathbf{S}$. The cross-modality encoder employs the Cross-Attention to Concatenation strategy, where we first perform two streams of cross-attention, and then the output is concatenated and processed by another Transformer to model the global context. 

Let $\mathbf{S}_{\mathrm{T}}$ and $\mathbf{S}_{\mathrm{F}}$ denote the transcript text and video frame embeddings respectively. Let $\mathbf{S}$ denote the V2LC output segment embedding produced by the multimodal interactions:
\begin{equation}
\left\{\begin{aligned}
\mathbf{S}_{\mathrm{T}} & \leftarrow \mathrm{Attention}\left({Q}_{\mathrm{t}}, {K}_{\mathrm{f}}, {V}_{\mathrm{f}}\right), \\
\mathbf{S}_{\mathrm{F}} & \leftarrow\mathrm{Attention}\left({Q}_{\mathrm{f}}, {K}_{\mathrm{t}}, {V}_{\mathrm{t}}\right), \\
\mathbf{S} & \leftarrow \mathrm{Mean}\left(\mathrm{Transformer}\left(\mathrm{Concat}\left(\mathbf{S}_{\mathrm{T}}, \mathbf{S}_{\mathrm{F}}\right)\right)\right) 
\end{aligned}\right.
\end{equation}

Then, we create the live comment feature embedding $\mathbf{C}$ %
by encoding a segment $s_i$'s corresponding live comments $c_i$ with a second text encoder. We perform contrastive learning by maximizing the cosine similarity of the video segment embedding ${\mathbf{S}}$ and live comment embedding ${\mathbf{C}}$. %

The goal of the V2LC encoder is to project each input video segment $s_i$ into the live comment representation space, so we freeze the live comment encoder and optimize only the video segment-to-live comment loss, training the V2LC encoder %
to match the representation of $s_i$'s accompanying live comments. %
The resulting training objective is: 
\begin{equation}
\mathcal{L}=-\frac{1}{N} \sum_{i=1}^N\left[\log \frac{e^{\left({\mathbf{S}}_i \cdot{\mathbf{C}}_i\right) * \tau}}{\sum_{j=1}^N e^{\left({\mathbf{S}}_i \cdot {\mathbf{C}}_j\right) * \tau}}\right]
\end{equation}
where $N$ is the number of video segment-live comment pairs in each batch, $({\mathbf{S}}_i \cdot {\mathbf{C}}_i)$ is the similarity (dot product) between video segment embedding ${\mathbf{S}}_i$ and live comment embedding ${\mathbf{C}}_i$, and $\tau$ is a learned temperature parameter. 

We use the V2LC encoder to produce synthetic live comment features to augment models for downstream tasks.

\subsection{Multi-label Objective}
The original CLIP architecture was designed primarily for single-label classification tasks. In our setting, there exists a many-to-many relationship, where a single video segment usually contains multiple live comments, and identical comments may appear in multiple video segments. To reconcile this discrepancy, we have modified the CLIP training framework to accommodate multiple labels: Given a batch of $N$ videos and $K$ live comments, the model is trained to predict which of the $N \times K$ possible (video, live comment) pairings across a batch actually co-occur.

An additional challenge arises when multiple similar comments from different video segments appear within the same batch. %
In our dataset, similar comments naturally occur more frequently than the more distinct content pairs seen in the image-text datasets used in CLIP. During pre-training, if similar comments from different video segments are sampled within the same batch, it might lead to confusion in the contrastive learning setup.
To mitigate this issue, we refine the training targets such that for a given video segment $s$, any live comment $c$ in the batch with a vector similarity exceeding a predefined threshold $\theta$ to any actual live comment of $s$ is also included in the target label set. %
Such comments are deemed correct, thereby allowing the model to learn a good live comment representation more efficiently.

\section{Downstream Fine-tuning}
\subsection{Multimodal Fusion Encoder}
Currently, in multimodal affective analysis, the text modality is commonly encoded with pre-trained language models like BERT \citep{devlin-etal-2019-bert}. %
In contrast, the acoustic and visual modalities have traditionally relied on hand-crafted feature extractors such as OpenSmile \citep{10.1145/1873951.1874246} for audio and OpenFace \citep{7477553} for facial expressions \citep{10.1145/3536221.3556630}.
These tools manually define the features to be extracted, which might not capture the full complexity of the data. However, recent %
work has introduced large-scale pre-trained encoders for these modalities which have yielded %
significant improvements %
in multimodal affective analysis. %
Therefore, we propose a downstream Multimodal Fusion Encoder that utilizes large-scale pre-trained encoders across all three modalities—text, acoustic, and visual—to allow more comprehensive %
modeling of %
affective tasks. %

We first extract features $f_m$ from each modality with pre-trained encoders, then we use a cross-modality encoder to fuse them.  
To accommodate three modalities, we construct the following cross-modality encoder:
\begin{equation}
\left\{\begin{aligned}
\mathbf{Z}_{\mathrm{t}} & \leftarrow \mathrm{Self Attn.}\left(\mathrm{Attention}\left({Q}_{\mathrm{t}}, {K}_{\mathrm{a+v}}, {V}_{\mathrm{a+v}}\right)\right), \\
\mathbf{Z}_{\mathrm{a}} & \leftarrow\mathrm{Self Attn.}\left(\mathrm{Attention}\left({Q}_{\mathrm{a}}, {K}_{\mathrm{v+t}}, {V}_{\mathrm{v+t}}\right)\right),
\\
\mathbf{Z}_{\mathrm{v}} & \leftarrow\mathrm{Self Attn.}\left(\mathrm{Attention}\left({Q}_{\mathrm{v}}, {K}_{\mathrm{t+a}}, {V}_{\mathrm{t+a}}\right)\right),
\\
\mathbf{Z} & \leftarrow \mathrm{Concat}\left(\mathrm{Mean}\left(\mathbf{Z}_{\mathrm{t}}\right), \mathrm{Mean}\left(\mathbf{Z}_{\mathrm{a}}\right), \mathrm{Mean}\left(\mathbf{Z}_{\mathrm{v}}\right)\right), 
\end{aligned}\right.
\end{equation}
where
\begin{equation}
\begin{split}
\begin{aligned}
\mathbf{Q}_{m 1} & =W_q \cdot f_{m 1} \\
\mathbf{K}_{m 2+m 3} & =W_k \cdot \mathrm{Concat}( f_{m 2}, f_{m 3}) \\
\mathbf{V}_{m 2+m 3} & =W_v \cdot \mathrm{Concat}( f_{m 2}, f_{m 3})
\end{aligned}
\end{split}
\end{equation}
Finally, we pass the concatenated features $\mathbf{Z}$ through feed-forward layers to produce our final prediction.

\subsection{Augmenting with Live Comment Features}

To enhance performance on downstream tasks, first, we apply self-attention to the synthetic live comment features produced by the V2LC encoder before mean-pooling. Then, we take the mean of these attentionally-pooled synthetic live comment features and concatenate it with $\mathbf{Z}$, our multimodal representation from our fusion encoder. These combined features are then processed through feed-forward layers. We hypothesize that this joint training enables rich integration of the emotional context captured by the synthetic live comment features from our V2LC encoder, improving their efficacy in downstream affective analysis.

\section{Experiments}
\subsection{Evaluation Datasets}
We evaluated our work on six widely used affective analysis datasets: Chinese Multimodal Sentiment Analysis Dataset (CH-SIMS) \citep{yu-etal-2020-ch}, Chinese Multimodal Sentiment Analysis Dataset v2.0 (CH-SIMS v2) \citep{10.1145/3536221.3556630}, Multimodal Opinion Sentiment and Emotion Intensity (MOSI) \citep{zadeh2016mosimultimodalcorpussentiment}, Multimodal Sentiment Analysis (MOSEI) \citep{bagher-zadeh-etal-2018-multimodal} for sentiment analysis, Multimodal EmotionLines Dataset (MELD) \citep{poria-etal-2019-meld} for emotion recognition, and Multimodal Sarcasm Detection Dataset (MUStARD) \citep{castro-etal-2019-towards} for sarcasm detection. Summaries of the datasets can be found in Table \ref{tab:down_dataset_stats}. %
\begin{table}
 \setlength{\tabcolsep}{1mm}
  \centering

\begin{tabular}{lcccc}
\toprule
Dataset & Lang &  \makecell{\#Total\\ segments} & \makecell{Avg \\duration (s)} & \makecell{Avg \\word count} \\
\cmidrule(r){1-1}\cmidrule(r){2-5}
CH-SIMS    & zh   &  2,282    &   3.67       &  15.8             \\
CH-SIMS v2 & zh   & 4,402    &   3.63       &  17             \\
MOSI    & en   & 2,199   &    4.2     &    12            \\
MOSEI   & en   & 22,856    &   7.28       &  19.7              \\
MELD    & en   & 13,708     &  3.59     & 8.0                \\
MUStARD & en   &  690   &    19.17       & 42.4        \\
\bottomrule
\end{tabular}
  \caption{
    Statistics of our evaluation datasets. 
  }
  \label{tab:down_dataset_stats}
\end{table}

CH-SIMS: This dataset contains short video segments from Chinese TV shows and movies, annotated with modality-specific and multimodal sentiment. We predict the multimodal sentiment labels.

CH-SIMS v2: An extension of the original CH-SIMS dataset, CH-SIMS v2 adds an additional $2,121$ video segments and improves the balance and diversity of the dataset.

MOSI: This dataset contains English movie review monologues from YouTube labeled with multimodal sentiment. 

MOSEI: An extension of MOSI, MOSEI expands its coverage to include more videos and topics.

MELD: %
Drawn from the TV show {\em Friends}, this dataset contains utterance-level labels chosen from $7$ emotions: anger, sadness, joy, neutral, fear, surprise, or disgust.

MUStARD: This dataset contains video clips from {\em Friends}, {\em The Golden Girls}, {\em Sarcasmaholics Anonymous}, and {\em The Big Bang Theory}. These audiovisual utterances, with their surrounding context, are annotated with sarcasm labels. 

\begin{table*}[t]
\centering
\begin{tabular}{@{}lcllllllllll}
\toprule
\multirow{2}{*}{\textbf{Models}} &\multirow{2}{*}{Modality}& \multicolumn{6}{c}{CH-SIMS v2}& \multicolumn{4}{c}{CH-SIMS }\\&&$\textrm {Acc}_{2}$ &  $\textrm {Acc}_{2}^{\textrm {w}}$& F1 & Corr & $\textrm{R}^2$ & MAE &$\textrm {Acc}_{2}$& F1 & Corr&MAE \\
\cmidrule(r){1-1}\cmidrule(r){2-2}\cmidrule(r){3-8}\cmidrule(r){9-12}
MulT&T+A+V&79.50& 69.61& 79.59& 70.32& 47.15& 31.7 &78.56&  79.66 &56.41&45.3\\

AV-MC  &T+A+V&   83.46 & 74.54& 83.52& 76.04 & 57.37& 28.6&-&-&-&- \\
VLP2MSA&T+V&-&-&-&-&-&-&79.43&79.26&60.38&41.29\\
MMML&T+A &-&-&-&-&-&-&82.93& 82.9& 73.26& 33.2\\
\cmidrule(r){1-1}\cmidrule(r){2-2}\cmidrule(r){3-8}\cmidrule(r){9-12}
Ours  &T+A+V &  86.94     & 76.70       &  86.88    &   82.32 & 66.71   &  26.6 &  83.16&83.16&74.32&32.8

\\
+ LC  &T+A+V & \textbf{90.12}\textsuperscript{*}  & \textbf{80.24}\textsuperscript{*} & \textbf{90.10}\textsuperscript{*}  & \textbf{86.14}\textsuperscript{*}  & \textbf{70.82}\textsuperscript{*}  &    \textbf{25.9}\textsuperscript{*}&\textbf{84.88}\textsuperscript{*}&\textbf{84.89}\textsuperscript{*}&\textbf{76.12}\textsuperscript{*}&\textbf{31.9}\textsuperscript{*} 

\\
\bottomrule

\end{tabular}
\caption{Results on CH-SIMS and CH-SIMS v2 for Chinese sentiment analysis. The performances of baseline models are shared by their authors. Our experimental results are averages across three random seeds. T, A, and V indicate text, acoustic, and visual modalities, respectively. ``Ours" denotes our vanilla Multimodal Fusion Encoder and ``+ LC" denotes our model augmented with synthetic live comment features. Asterisks indicate t-test $p\,$\textless$\,0.05$ when compared with our vanilla Multimodal Fusion Encoder.}
\label{Table_CSA}
\end{table*}

\begin{table*}[t]
\centering
\begin{tabular}{@{}lcllllllllll}
\toprule
\multirow{2}{*}{\textbf{Models}}&\multirow{2}{*}{Modality} & \multicolumn{5}{c}{MOSI}& \multicolumn{5}{c}{MOSEI}\\&&$\textrm{Acc}_{2}$ & $\textrm{Acc}_{7}$ & F1 & Corr & MAE &$\textrm{Acc}_{2}$ &$\textrm{Acc}_{7}$ &F1 & Corr &  MAE  \\
\cmidrule(r){1-1}\cmidrule(r){2-2}\cmidrule(r){3-7}\cmidrule(r){8-12}
MulT &T+A+V& 84.10& -& 83.90& 71.1&  86.1&82.5& -&82.3  &71.3 &58.0\\
UniMSE&T+A+V&86.9&48.68&86.42&80.9&69.1&\textbf{87.50}&54.39&\textbf{87.46}&77.3 &52.3\\
VLP2MSA&T+V&86.28&-&86.26&81.3&69.6&85.97&-&85.89&77.0&53.5\\
MMML&T+A& 88.16& 48.25& 88.15 &83.8& 64.29 &86.73& 54.95&86.49&79.08&51.74\\
\cmidrule(r){1-1}\cmidrule(r){2-2}\cmidrule(r){3-7}\cmidrule(r){8-12}
Ours   &T+A+V& 88.26 &49.62  & 88.17 & 84.10& 63.4& 86.24 & 54.95 &86.21  & 79.10 & 51.6  \\
+ LC   &T+A+V   &  \textbf{89.02}\textsuperscript{*}&\textbf{50.14}\textsuperscript{*}&\textbf{89.06}\textsuperscript{*}&\textbf{84.50}&\textbf{63.2}&87.15\textsuperscript{*}& \textbf{55.15}& 87.12\textsuperscript{*} &\textbf{80.22}\textsuperscript{*} & \textbf{51.3}

\\
\bottomrule
\end{tabular}
\caption{Results on MOSI and MOSEI for English sentiment analysis.  %
}
\label{Table_ESA}
\end{table*}

\begin{table}
\centering
\begin{tabular}{lclllll}
\toprule
\multirow{2}{*}{\textbf{Models}}&\multirow{2}{*}{Modality}  &\multicolumn{1}{c}{MELD}& \multicolumn{3}{c}{MUStARD}\\&&$\textrm{F1}_{\textrm {w}}$ &$\textrm{P}_{\textrm {w}}$&$\textrm{R}_{\textrm {w}}$&$\textrm{F1}_{\textrm{w}}$ \\
\cmidrule(r){1-1}\cmidrule(r){2-2}\cmidrule(r){3-3}\cmidrule(r){4-6}
SVM &T+A&-&65.2& 62.9& 63.0\\
UniMSE &T+A+V&65.51&-&-&-\\
\cmidrule(r){1-1}\cmidrule(r){2-2}\cmidrule(r){3-3}\cmidrule(r){4-6}
Ours & T+A+V &      66.73  &   68.2&67.4&67.5    \\
+ LC  & T+A+V &      \textbf{69.62}\textsuperscript{*}&   \textbf{70.6}\textsuperscript{*}&\textbf{70.5}\textsuperscript{*}  &\textbf{70.5}\textsuperscript{*}
\\
\bottomrule

\end{tabular}
\caption{Results on MELD for emotion recognition and MUStARD for sarcasm detection.  %
}
\label{Table_MM}
\end{table}
\subsection{Implementation}
\subsubsection{Data Processing}
For pre-training, we use a segment length $\sigma$ of $8$ seconds, %
balancing between required context and the length of downstream datasets. We sample $8$ frames uniformly from each segment.
We filter out comments that contain no meaningful information such as comments that are too short (less than $2$ characters).  
For user generated videos, we trim the first and last $15$ seconds as they tend to include repetitive comments such as greetings and farewells.
For movies, we trim the first and last $5$ minutes, and for TV shows, we trim the start and end of each show. 

Segments containing fewer than $5$ live comments are excluded from pre-training to allow efficient GPU batching. %
For each epoch, we randomly select $5$ live comments per segment so that a batch with $N$ samples has $K=5N$ comments. 
We sample $10\%$ of our data for validation.

\subsubsection{Implementation Details}

We pre-train two variants of our V2LC encoder: one that uses only Chinese text transcripts and another that uses English text transcripts when available. For the Chinese-only variant, we use \texttt{Chinese-RoBERTa-wwm-ext} \citep{cui-etal-2020-revisiting} as our transcript encoder; for the bilingual variant, we use \texttt{XLM-RoBERTa-base} \citep{conneau2019unsupervised}. For encoding live comments, We use \texttt{Chinese-RoBERTa-wwm-ext}.

For downstream fine-tuning, we use pre-trained encoders tailored to each modality and language. For the text modality, we use \texttt{Chinese-RoBERTa-wwm-ext} for Chinese datasets and \texttt{RoBERTa-base} for English datasets \citep{liu2019roberta}. For the acoustic modality, we adopt \texttt{Chinese-HuBERT-base} to encode Chinese data and \texttt{Data2Vec-audio-base} for English data. For visual features, we process all data using  \texttt{TimeSformer-base}.

Other details are specified in the Appendix.%

\subsection{Baseline Models}
We compare our model against multiple competitive baselines:

Multimodal Transformer (MulT) \citep{tsai-etal-2019-multimodal}: Chosen for its pioneering approach of applying attention %
across modalities,
MulT enables dynamic adaptation between modalities at different time steps, addressing the challenges posed by unaligned multimodal data. 

Support Vector Machine (SVM): Known for its robust performance on small-sized datasets, SVM can, at times, surpass neural models. Following \citet{castro-etal-2019-towards}, features %
are concatenated via early fusion and fed into an SVM classifier.

Acoustic Visual Mixup Consistent framework (AV-MC) \citep{10.1145/3536221.3556630}: The current SOTA for CH-SIMS v2, AV-MC introduces a modality mixup module as data augmentation, which mixes the acoustic and visual modalities from different videos to enhance performance.

Multimodal Sentiment Knowledge-sharing Framework (UniMSE) \citep{hu-etal-2022-unimse}: Chosen for its strong performance on MOSEI and MELD, this model proposes a knowledge-sharing framework that unifies two affective analysis tasks: multimodal sentiment analysis and emotion recognition in conversation, demonstrating the effectiveness of utilizing complementary knowledge in affective tasks.

Vision-Language Pre-Training To Multimodal Sentiment Analysis (VLP2MSA) \citep{YI2024111136}: Chosen for its use of pre-trained vision models, VLP2MSA extracts spatio-temporal features from sparsely sampled video frames, offering advantages over traditional visual feature extraction. It captures not only facial expressions but also body movements, providing a more comprehensive analysis of visual information. 

Multimodal Multi-loss Fusion Network (MMML) \citep{wu2024multimodal}: The current SOTA for CH-SIMS and MOSI, MMML utilizes pre-trained acoustic models as feature extractors for the acoustic modality.

\subsection{Metrics}
We evaluate model performance using standard metrics for each task as established in the literature. Further details can be found in the Appendix.
\subsection{Results}

We present our experimental results on sentiment analysis in Tables \ref{Table_CSA} and \ref{Table_ESA} and emotion recognition and sarcasm detection in Table \ref{Table_MM}.  First, we note that our vanilla Multimodal Fusion Encoder before live comment feature augmentation (as indicated by ``Ours") outperforms all of our baselines in nearly every metric across all tasks.  This demonstrates the strength of utilizing pre-trained encoders for all modalities in affective analysis and establishes a new SOTA on its own. 

When we treat this vanilla Multimodal Fusion Encoder as a strong baseline, we see that the addition of synthetic live comment features (as indicated by ``+ LC") yields additional gains, outperforming the vanilla Multimodal Fusion Encoder (and our other baselines) by significant margins. 

For Chinese sentiment analysis, we observe a  $3.18$-point increase in accuracy and a $3.82$-point improvement in Pearson Correlation (Corr) on CH-SIMS v2, and a $0.9$-point decrease in Mean Absolute Error (MAE) on CH-SIMS. 

For English, in emotion recognition, we see a $2.89$-point increase in weighted F1-score on MELD. In sarcasm detection, we see a $3.0$-point increase in weighted F1-score on MuSTARD.  And in sentiment analysis, we also see improvements, though more modest. This is likely due to the genre shift from our pre-training dataset to the English sentiment analysis benchmarks. As noted in the dataset section, MOSI and MOSEI primarily contain YouTube monologues featuring one speaker which differs significantly from the conversational content in our collected pre-training corpus.

\subsection{Affective Features Comparison Study}
\begin{table}
\centering
\begin{tabular}{lccccc}
\toprule
\multirow{2}{*}{Features} & \multicolumn{2}{c}{Logistic Reg.} & \multicolumn{3}{c}{w/ Multi. Fusion Enc.} \\
&$\textrm{Acc}_{2}$&F1&$\textrm{Acc}_{2}$&Corr&MAE\\
\cmidrule(r){1-1}\cmidrule(r){2-3}\cmidrule(r){4-6}
Erlangshen & 72.89& 72.83  & 86.28  & 82.21& 27.3                  \\
COLD. & 75.63 &     75.62    & 87.27 & 83.72 &26.4            \\
VIT-face&78.70&78.55  &  86.39  &82.69&26.6                    \\
\cmidrule(r){1-1}\cmidrule(r){2-3}\cmidrule(r){4-6}
Ours (LC)  & 73.66&73.68  & 90.12 &86.14  &25.9                  \\
\bottomrule
\end{tabular}
\caption{Results of our feature comparison study. %
}
\label{tab_fcs}
\end{table}

\subsubsection{Affective Feature Baseline Models}

We compare our synthetic live comment features with features from models trained for related affective analysis tasks:

\textit{Text-based Chinese sentiment} features from Erlangshen-Roberta-sentiment (Erlangshen) \citep{fengshenbang}: representations from a text sentiment analysis model (\texttt{Chinese-RoBERTa-wwm-ext} fine-tuned on $227, 347$ sentiment-labeled texts from $8$ datasets).

\textit{Text-based Chinese offensive language} features from COLDETECTOR \citep{deng-etal-2022-cold}: representations from an offensive language detection model (\texttt{Chinese-RoBERTa-wwm-ext} fine-tuned on $37,480$ sentences annotated with binary offense labels from the COLDataset).

\textit{Image-based facial emotion recognition} features from Vit-face-expression (VIT-face) \citep{todor_pakov_2024}: representations from a facial emotion recognition model (Vision Transformer \citep{dosovitskiy2020image} fine-tuned on $35,887$ faces from the FER2013 dataset, annotated with one of seven emotions: Anger, Disgust, Fear, Happiness, Sadness, Surprise, and Neutral).

\subsubsection{Affective Feature Experiments}

We evaluate the effectiveness of our synthetic live comment features through controlled experimentation in two settings.  In the first, we train a logistic regression classifier on individual features to measure how well each feature discriminates our evaluation data.  In the second, we augment our vanilla Multimodal Fusion Encoder with individual features to evaluate how well each feature can be leveraged by more expressive multimodal models.  For these experiments, we focus on Chinese sentiment analysis with CH-SIMS v2 due to the diversity of its content and the amount of data available for fine-tuning.  Additional experimental details can be found in the Appendix.  We present the results of these experiments in Table \ref{tab_fcs}.

\subsubsection{Affective Feature Comparison Results}

When using a simple logistic regression classifier, our synthetic live comment features outperform text-based Chinese sentiment from the Erlangshen even though that model was trained on a large quantity of task-specific sentiment data.  Additionally, though less performant, our features are competitive with our other baseline features from models that benefit from supervised training on related affective tasks.

The strength of our synthetic live comment features becomes apparent when we move to our Multimodal Fusion Encoder. In this setting, they outperform the baselines by wide margins across all metrics considered.  We hypothesize that though our synthetic live comment features are less discriminative in a linear model for a specific affective task (like sentiment analysis), they encode rich information that can be adapted to any affective task by the more expressive multimodal models that have become the standard in AI today.  Our experimental results speak to this effectiveness.

\section{Conclusion}

In this work, we demonstrate the efficacy of synthetic live comment features in multimodal affective analysis. By constructing the LCAffect dataset and training a multimodal encoder on it, we enable the inference of these synthetic features for any video content. Our enhanced multimodal model, augmented with synthetic live comment features, achieves new SOTA results in sentiment analysis, emotion recognition, and sarcasm detection across both English and Chinese content. This advance confirms that synthetic live comment features can effectively capture a broad spectrum of affective states, opening new avenues across diverse dimensions of affective analysis.

\bibliography{aaai25}

\end{document}